\pdfoutput=1

\documentclass[11pt]{article}
\usepackage{coling2020}
\usepackage{times}
\usepackage{url}
\usepackage{latexsym}

\usepackage[T1]{fontenc}
\usepackage[utf8]{inputenc}
\usepackage[textsize=tiny]{todonotes}
\usepackage{graphicx}
\usepackage{booktabs}
\usepackage{latexsym}

\usepackage{footnote}
\makesavenoteenv{tabular}
\makesavenoteenv{table}
\makesavenoteenv{table*}

\newcommand\minput[1]{%
  \input{#1}%
  \ifhmode\ifnum\lastnodetype=11 \unskip\fi\fi}

\usepackage{hyperref}
\usepackage{xstring}

\newcommand{\code}[1]{\texttt{#1}}
\newcommand{\noqa}[1]{}
\newcommand{\noqall}[1]{}

\usepackage{amsmath}

\usepackage{multirow}
\usepackage{longtable}

\newcommand{\entity}[1]{{\small{\texttt{#1}}}}

\newcommand\Mark[1]{\textsuperscript{#1}}

\title{Kleister: A novel task for Information Extraction\\involving Long Documents with Complex Layout}

\author{
 \parbox{\linewidth}{\centering\normalfont\large
    Filip Graliński\Mark{1,2}, Tomasz Stanisławek\Mark{1,4}, Anna Wróblewska\Mark{1,4},\\Dawid Lipiński\Mark{1}, Agnieszka Kaliska\Mark{1,2}, Paulina Rosalska\Mark{1,3},\\Bartosz Topolski\Mark{1}, Przemysław Biecek\Mark{4,5}%
  }\\
 {%
    \Mark{1}\small{Applica.ai, 15 Zajęcza, Warsaw, 00351, Poland, \texttt{firstname.lastname@applica.ai}}
    } \\
 {%
    \Mark{2}\small{Adam Mickiewicz University, 1 Wieniawskiego, Poznan, 61712, Poland, \texttt{firstname.lastname@amu.edu.pl}}
    } \\
 {%
    \Mark{3}\small{Nicolaus Copernicus University, 11 Gagarina, Torun, 87100, Poland, \texttt{firstname.lastname@umk.pl}}
 } \\
 {%
    \Mark{4}\small{Warsaw University of Technology, Koszykowa 75, Warsaw, Poland, \texttt{firstnameletter.lastname@pw.edu.pl}}
    } \\
{%
    \Mark{5}\small{Samsung R\&D Institute Poland, Plac Europejski 1, Warsaw, Poland,  \texttt{firstnameletter.lastname@samsung.com}}
    } \\
}

\begin{document}
\maketitle
\begin{abstract}
State-of-the-art solutions for Natural Language Processing (NLP) are able to capture a~broad range of contexts, like the sentence-level context or document-level context for short documents. But these solutions are still struggling when it comes to longer, real-world documents with the information encoded in the spatial structure of the document, such as page elements like tables, forms, headers, openings or footers; complex page layout or presence of multiple pages.

To encourage progress on deeper and more complex Information Extraction (IE) we introduce a~new task (named \noqa{spell-Kleister}\emph{Kleister}) with two new datasets. Utilizing both textual and structural layout features, an~NLP system must find the most important information, about various types of entities, in long formal documents. We propose \textit{Pipeline} method as a text-only baseline with different Named Entity Recognition architectures (Flair, BERT, RoBERTa). Moreover, we checked the most popular PDF processing tools for text extraction (\noqa{spell-pdf2djvu}pdf2djvu, Tesseract and Textract) in order to analyze behavior of IE system in presence of errors introduced by these tools.

\end{abstract}

\noqall{spell-pdf2djvu}
\noqall{spell-djvu2hocr}

\section{Introduction}

Information Extraction\noqall{spell-Kleister} (IE) requires quick but careful skimming through the whole document. We often have to not only search for pieces of information, but also to generate final output for specific entity type (e.g. aggregate multiple occurrences of organization names into one). In practice, this means that the results should be presented in an appropriate form (e.g.~data points such as addresses normalized to a standard form). It should also be explained why certain information has been correlated. This may take the form of an indication in the input text. The process can be tedious and difficult for humans to do. Thus, we need automated systems to cope with multiple documents and to extract the required information in a~simple and~efficient way.

However, the disparity between what can be done with the state of the art in IE and what is required by real-world business use cases is still large.
From the point of view of business users, systems that automatically gather information about individuals, their roles, significant dates, addresses, and amounts from invoices, companies reports and contracts, would be useful~\cite{holt-chisholm-2018-extracting,DBLP-journals-corr-abs-1809-08799,DBLP-conf-fedcsis-WroblewskaSPG18,DBLP-journals-corr-abs-1906-02427}. Furthermore, the systems should be reliable and should reliably assess their own certainty about extracted entities.

However, as far as the state of the art is concerned, there are many machine learning models which must be trained for general named entities to be robust~\cite{DBLP-journals-corr-abs-1802-05365,akbik-etal-2018-contextual,DBLP-journals-corr-abs-1810-04805}. To further increase training efficiency, we can use the documents of a~previously defined layout, so that the models could learn how to extract a~particular piece of~information
\cite{Xiaohui2019b,denk2019,Xiaojing2019,Sarkhel2019}.
On the other hand, more general extractors are still needed to deal with a~variety of information.

In this paper, we describe two novel datasets for Information Extraction from long documents with complex layouts. We will begin by explaining the need for the dataset that would contain authentic scenarios to provide a~review of similar tasks and datasets in the next step (Section~\ref{sec:review}). Then, we describe characteristics of datasets in details  (Section~\ref{sec:dataset}). Subsequently, we describe baseline methods (using only textual information, without relying directly on 2D information) and their results (with different PDF processing tools) applied to cope with the task with the \textit{Pipeline} approach described in Section~\ref{sec:baselines}. Finally, we discuss challenges in the process to extract the proper entities (Section~\ref{sec:discussion}).

\begin{figure*}[ht!]%
\begin{center}
\includegraphics[width=16cm]{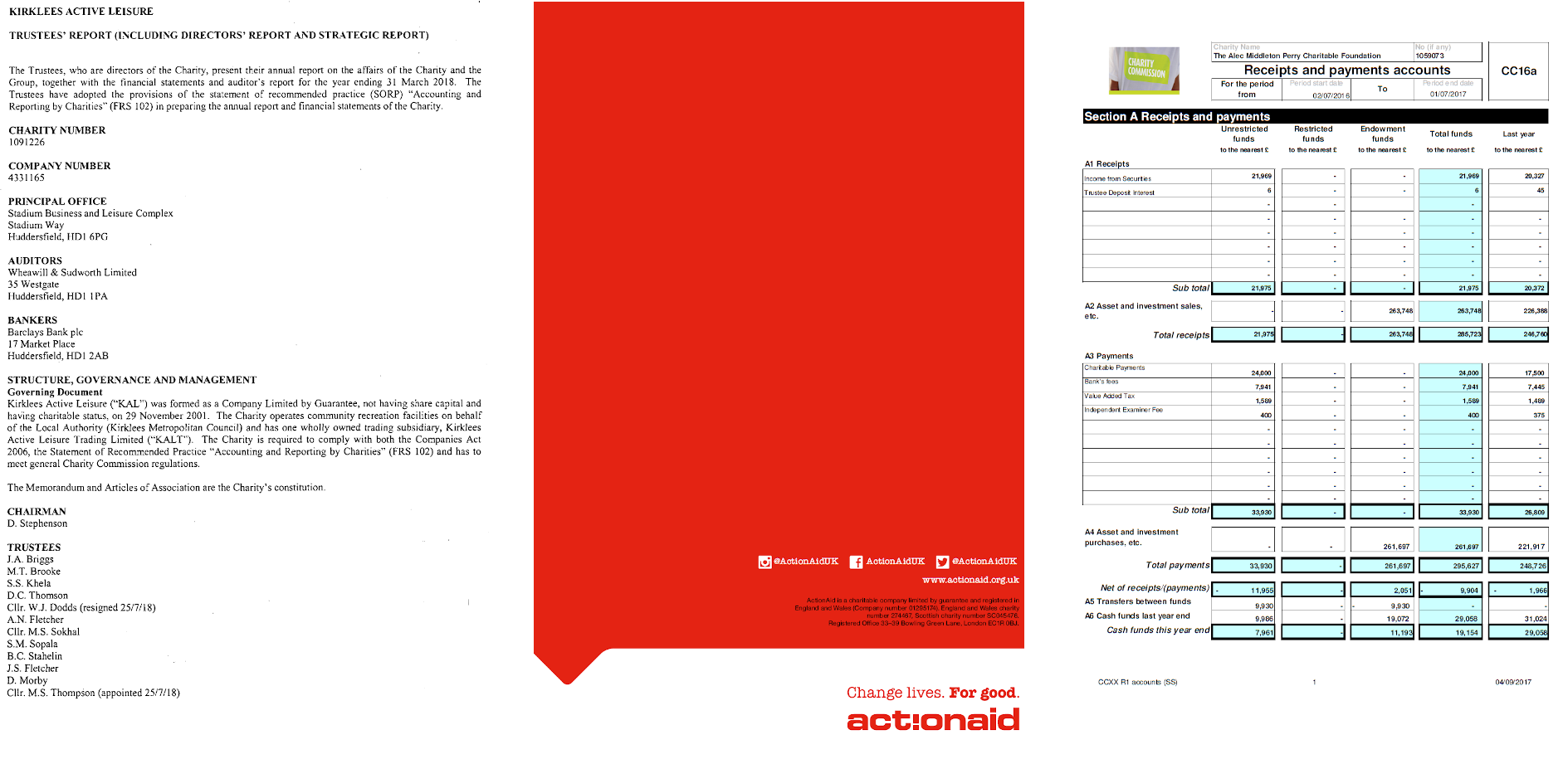}
\caption{Three different examples of layouts from the \emph{Kleister-charity} dataset.\label{fig:charity_layout}}
\end{center}
\end{figure*}

\section{Review of Information Extraction Datasets}
\label{sec:review}

Our main idea for preparing a~new dataset was to develop a~strategy to deal with the main challenge we face in business conditions, which means overcoming such difficulties as: complex layout, specific business logic (the way the content is formulated), OCR quality, document-level extraction and normalization.

The most similar dataset to our approach regarding the NLP field is the WikiReading dataset and related challenges~\cite{hewlett-etal-2016-wikireading}. This dataset is a~large-scale natural language understanding task with 18 million entities and 4.7 million documents. The goal of the task is to predict textual values from the structured knowledge base, Wikidata, by reading the text of the corresponding Wikipedia articles. Some entities can be extracted from the given text, but some of them have to be inferred. Thus, similarly as in our assumptions, the task contains a~rich variety of challenging extraction sub-tasks and is also well-suited for end-to-end models. In both sets there are also challenges with output data normalization, e.g.~dates, names.

However, our datasets are even more difficult to process, because they comprise documents with complex layout, noisy OCR-ed input (made by an Optical Character Recognition system) and they are much longer than an average Wikipedia article. These are the main issues that distinguish it from WikiReading and which justify why our task is not only about understanding the language.

A~list of challenges similar to some degree to our goal is also available at the International Conference on Document Analysis and Recognition ICDAR 2019\footnote{\url{http://icdar2019.org/competitions-2/}}. However, the authors focus mainly on understanding tables and a limited range of document layouts, not extracting particular information from the data. There is a~dataset called Form Understanding in Noisy Scanned Documents \noqa{spell-FUNSD}(FUNSD)~\cite{jaume2019}. FUNSD aims at extracting and structuring the textual content of forms. Unfortunately, the dataset comprises only 200 scanned and annotated forms and the annotations are too general, i.e.~question, answer, header.

Another interesting dataset from ICDAR 2019 is a~set of scanned receipts. The authors prepared 1000 whole scanned receipt images with annotations of company name, date, address and the total payment amount\footnote{\url{https://rrc.cvc.uab.es/?ch=13}}. Of course, receipts are short documents, and have quite a~uniform layout and information structure (they start with the company name, date and invoice number etc.).

Finally, there are datasets with the information extraction task based on invoices, which are not publicly available to the community~\cite{holt-chisholm-2018-extracting,DBLP-journals-corr-abs-1809-08799}. Documents of this kind contain common entities like ‘Invoice date’, ‘Invoice number’, ‘Net amount’ and ‘Vendor Name’ which are extracted using a~combination of Natural Language Processing and Computer Vision techniques. This is because spatial information is important to properly understand such documents. However, since they are usually short, there is rather no repetition of the same information, so there is no need to understand the context.

\section{Kleister: New Datasets}
\label{sec:dataset}

The main goal of the gathered datasets introduced in this paper is to emphasize business value and focus more on problems related to layout analysis and Information Extraction, as well as Natural Language Understanding (several entities should be inferred from the whole document context). Thus, it can be performed as well as an end-to-end task that can be used in real life use cases for robotic process automation of information extraction from documents of complex layout.

We collected datasets of long formal documents that are US non-disclosure agreements (\emph{Kleister-NDA}) and annual financial reports of charitable foundations in the UK (\emph{Kleister-Charity}).
The datasets (training and development sets as well as the input to
the main test set)%
\footnote{We are planning to set up a shared-task platform where submissions could be evaluated for the test set as well.}
are available at \url{https://github.com/applicaai/kleister-nda}
and \url{https://github.com/applicaai/kleister-charity}.

\emph{Kleister} datasets have a~multi-modal input (i.e.~text versions were obtained from OCR-ed noisy documents, some of which contain illustrations and some were scans) and a~list of entities to be found. The list of reference findings is not indicated in the input documents. This is not a~NER task, in which we would be interested in determining where a~given piece of information or entity is in the text. We are interested in the information itself. Moreover, we assume that in our datasets some documents may be missing some entities, or some entities may have more than one gold value. The input of the dataset comprises: PDF files and text versions of the documents (we used most popular tools for extraction text from PDF files).

These two datasets have been gathered in different ways because of their repository structures. Also, the reasons why they were published on the Internet were different. The most important difference between them is that the NDA dataset was born-digital but the Charity dataset needed to be OCR-ed.
The detailed information about aforementioned open datasets (which are the most popular ones in the domain) and our \emph{Kleister} datasets are presented in Table~\ref{tab:dataset_summary}.

\subsection{NDA Dataset}
\label{sec:NDA_dataset}

The\noqa{grammar-ENGLISH_WORD_REPEAT_BEGINNING_RULE} NDA Dataset contains Non-disclosure Agreements, also known as Confidential Agreements. They are legally binding contracts between two or more parties, through which the parties agree not to disclose information covered by the agreement. The NDAs were collected from the Electronic Data Gathering, Analysis and Retrieval system (EDGAR) via Google search engine. Then, a~list of entities was established (see Table~\ref{tab:entities}) and documents were manually annotated by a~team of linguists, which ensured excellent quality of the annotation ($\kappa=0.971$\footnote{This is the average result for all entities — Cohen's kappa coefficient was calculated for each entities on the basis of double annotation of 100 random documents from the entire NDA Dataset. A~detailed description of the data collection method and annotation procedure can be found in the Appendix.}).

In Figure~\ref{fig:datasets_problems}, there is an example of a problematic entity in a Non-Disclosure Agreement: effective date. It is the date on which the contract enters into force. In general, it coincides with the date of the contract or the date it was signed. It happens, however, that these dates are different and then the date of entry into force of the contract is specially marked, e.g. as ‘Effective date’. However, none of dates in the figure is specified in this way. Most NDAs contain a special clause that indicates the date of entry into force of the contract. Usually it is immediately before the signatures of the parties. In this case, the correct answer is November 20, 2008, because in this agreement there is a clause: ‘IN WITNESS WHEREOF, he parties hereto have executed this Agreement on the date first written above.’

\begin{table*}[ht!]
\caption{Summary of the existing datasets and the \emph{Kleister} sets.\label{tab:dataset_summary}}
\centering
{\small{
\begin{tabular}{p{2.1cm}|p{1.7cm}p{2.6cm}p{1.6cm}p{1.6cm}|p{1.3cm}p{2cm}}\noqa{grammar-UNIT_SPACE}\noqa{latex-44}
\toprule
Dataset name & CoNLL 2003 & WikiReading & \noqa{spell-FUNSD}FUNSD ICDAR 2019 & SROIE\noqa{spell-SROIE} ICDAR 2019 & Kleister-NDA & Kleister-Charity\\
\midrule
Source & Reuters news & WikiData/ Wikipedia & scanned forms & scanned receipts & EDGAR & UK Charity Commission \\
Annotation & manual & automatic & manual & manual & manual & semi-automatic \\
Documents & 1 393 & 4.7M\noqa{spell-7M} & 199 & 973 & 540 & 2 778  \\
Entities & 35 089 & 18M\noqa{spell-18M} & 9 743 & 3 892 & 2 160 & 21 612\\
\midrule
train docs & 946 & 16.03M\noqa{spell-03M} & 149 & 626 & 254  & 1 729 \\
dev\noqa{spell-dev}\noqa{grammar-CD_NN}  docs &  216 & 1.89M\noqa{spell-89M} & — & — & 83  & 440 \\
test docs & 231 & 0.95M\noqa{spell-95M} & 50 & 347 & 203 & 609 \\
\midrule
Entity classes & 4 & 867 (top 20 cover 75\%) & 3 & 4 & 4 & 8 \\
\midrule
Mean pages/doc & — & 1/Wikipedia article & 1 & 1/receipt & 5.98 & 22.19\\
Mean words/doc & 216.4 & 489.2 & 158.2 & 45 & 2540 & 5149 \\
Mean entities/doc & 25.2 & 5.31 & 49.0 & 4 & 4.0 & 7.8 \\
\midrule
Complex layout & N & N & Y & Y & Y/N & Y/N \\
\bottomrule
\end{tabular}
}}

\setlength{\tabcolsep}{4pt}
\caption{Summary of the entities in the NDA and Charity datasets.\label{tab:entities}}
\centering
\small{
\begin{tabular}{p{4,0cm}p{8,0cm}ll}
\toprule
 Entities & Description & Total & \% all entities \\
 \midrule
 \multicolumn{4}{c}{\emph{NDA} dataset} \\
 \midrule
 \entity{party} & parties appearing in the agreement (each of them is treated as a~separate entity) & 1035 & 47.9 \\
  \entity{jurisdiction} & state or country whose law governs the agreement & 531 & 24.6  \\
 \entity{effective\_date} & date on which the contract becomes legally binding & 400 & 18.5 \\
 \entity{term} & duration of the agreement & 194 & 9.0 \\
 \midrule
 \multicolumn{4}{c}{\emph{Charity} dataset} \\
 \midrule
 \entity{address\_\_post\_town} & post town\noqa{grammar-EN_COMPOUNDS} (part of a~charity address) & 2692 & 12.5 \\
 \entity{address\_\_postcode} & postcode (part of a~charity address) & 2717 & 12.6 \\
 \entity{address\_\_street\_line} & street with the house number (part of a~charity address) & 2414 & 11.1 \\
 \entity{charity\_name} & name of the charitable organization & 2778 & 12.9 \\
 \entity{charity\_number} & identification number in the charity register & 2763 & 12.8 \\
 \entity{report\_date} & date of reporting & 2776 & 12.8 \\
 \entity{income\_annually} & annual income in British pounds (GBP) & 2741 & 12.7 \\
 \entity{spending\_annually} & annual spending in British pounds (GBP) & 2731 & 12.6 \\
 \bottomrule
\end{tabular}
}
\end{table*}

\begin{figure*}[hbt!]
\begin{tabular}{p{16,0cm}}\noqa{grammar-UNIT_SPACE}\noqa{latex-OVERFULL}
\begin{center}
 \includegraphics[width=12.0cm]{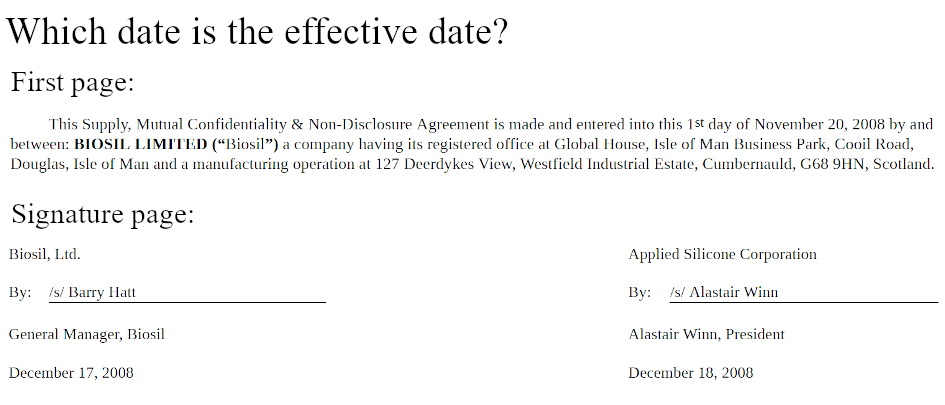} \includegraphics[width=14.0cm]{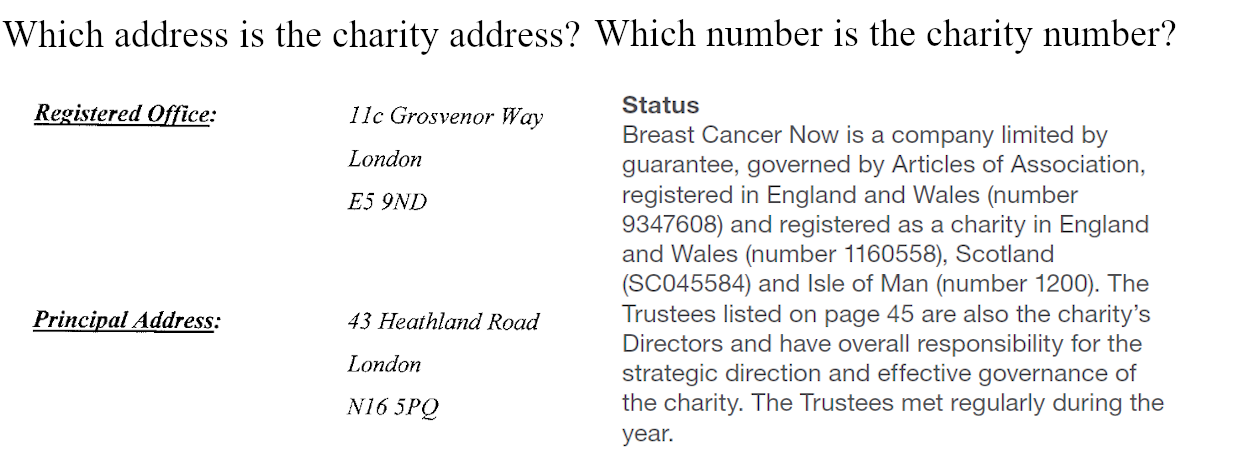}
\end{center}
\end{tabular}
\caption{Examples of problematic entities in documents from the Kleister-NDA and Charity datasets.\label{fig:datasets_problems}}
\end{figure*}

\subsection{Charity Dataset}
\label{sec:charity_dataset}

The Charity dataset consists of annual financial reports that all charities registered in England and Wales are required to submit to the Charity Commission for England and Wales. Then, the Commission makes them publicly available via its website.\footnote{\url{https://apps.charitycommission.gov.uk/showcharity/registerofcharities/RegisterHomePage.aspx}} Charity reports were collected from the UK Charity Commission website, just like annotations to these documents. The entity list was established on the basis of information that we were able to automatically obtain from the tables on the page describing the content of the reports\footnote{A~detailed description of the data collection method can be found in the Appendix.} (see Table~\ref{tab:entities}).%

The quality of automatically obtained entities was checked by a~team of annotators based on 100~random reports. After analyzing these documents, the following annotations were corrected: the names of the organizations (normalization of \textit{Ltd.}) and amounts (we fixed entities by adding a~decimal part of the value) in a~part of the development set and in the whole test set (development and test sets are important in context of measuring model actual performance). Then we repeated the annotation check based on 200~random documents from train and development sets (we assume that the annotation of the test set is excellent—$\kappa=0.848$\footnote{Cohen's kappa coefficient was calculated on the basis of double annotation of 100 random documents from the set test.}). Our preliminary and final results of the quality control procedure are presented in Table~\ref{tab:entities}. The results for the \textit{train} dataset are definitely lower, but at the same time this set is four times larger than the two others and, unlike them, only a~small part of it was manually annotated.

\begin{table}[htbp]
\caption{Results of the manual verification of Charity dataset.\label{tab:charity_data_verification}}
\centering
\small{
\begin{tabular}{p{4cm}p{3.9cm}p{1.2cm}p{1.2cm}p{1.2cm}}\noqa{grammar-UNIT_SPACE}
\toprule
 Entities & {Correct initial annotations[\%]} & \multicolumn{3}{c}{Correct final annotations[\%]}  \\
 & entire dataset & train & dev\noqa{spell-dev} & test ($\kappa$) \\
 \midrule
 \entity{address} (as a whole) & 23 & 55 & 93 & 0.920\\
 \entity{address\_\_post\_town} & — & 83 & 99 & 0.889\\
 \entity{address\_\_postcode} & — & 78 & 98 & 1.000\\
 \entity{address\_street\_line} & — & 67 & 93 & 0.871\\
 \entity{charity\_name} & 86 & 81 & 92 & 0.904\\
 \entity{charity\_number} & 99 & 95 & 100 & 0.492\\
 \entity{charity\_date} & 99 & 98 & 100 & 1.000\\
 \entity{income\_annually} & 82 & 90 & 91 & 0.906\\
 \entity{spending\_annually} & 78 & 86 & 92 & 0.725\\
\bottomrule
\end{tabular}
}
\end{table}

Figure~\ref{fig:datasets_problems} shows problems with two entities in reports of the charitable organization: charity address and number. Both can co-occur in many variants for the same organization and in the same document. In these cases, it was necessary to refer to the business logic, so the correct answers are “Registered address” and charity number for England and Wales.

\section{Baseline models}
\label{sec:baselines}

\begin{figure*}[ht!]
\begin{center}
\includegraphics[width=1\textwidth]{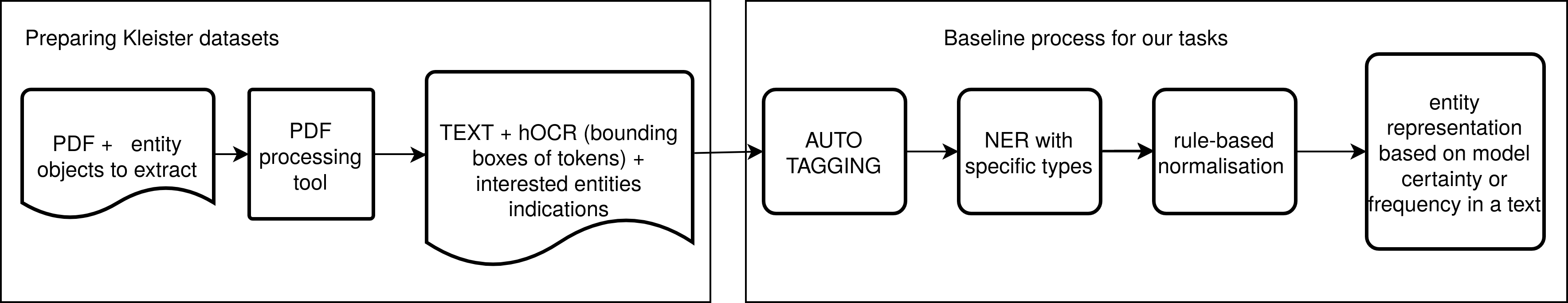}
\caption{Our process of preparing Kleister datasets and training baselines. Initially, we gathered PDF documents and required entities’ values; an~important part of the process is the OCR. Then,based on only textual data we prepare pipeline solutions. The pipeline process is illustrated in the second frame and consists of the following stages: auto-tagging, standard NER, text normalization, final selection of the values of entities.\label{fig:data-baselines}}
\end{center}
\end{figure*}

Kleister datasets for information extraction are challenging tasks and do not exactly match any existing solutions in the current NLP world. In this paper, our aim is to produce strong baselines based on text treated as a~sequence, without using additional spatial information. We propose \textit{Pipeline} technique to solve extraction problems. Our baseline \textit{Pipeline} method is a~chain of processes with a~named entity recognition (NER) model as a crucial one to indicate a~given entity in the text, then to normalize the entities to canonical forms and finally to aggregate all results into one, adequate to the given entity type. Contextual String Embeddings (“Flair”)~\cite{akbik-etal-2018-contextual},  BERT-base~\cite{DBLP-journals-corr-abs-1810-04805} and RoBERTa-base~\cite{Liu2019RoBERTaAR} models are used for this. Moreover, we tried different PDF processing tools for text extraction from PDF documents to check importance of text quality to final score of the system.

\subsection{Pipeline}
\label{sec:baseline_pipeline}

The core idea of this method is to select specific parts of the text in a~document that denote the objects that we are looking for. The whole process is presented in Fig.~\ref{fig:data-baselines} with the following stages:
\begin{enumerate}
    \item \textbf{Auto-tagging}: this stage involves extracting all the fragments that refer to the same or different entities by using sets of regular expressions combined with a~gold-standard value for each general entity type (date, organization, amount, etc.), e.g.~when we try to detect a~{\entity{report\_date}} entity, we must handle different date formats: `November 29, 2019’, `11/29/19’ or \noqa{latex-8}`11-29-2019’. This step is performed only during training (to get data on which a~NER model can be trained).
    \item \textbf{Named Entity Recognition}: using the auto-tagged dataset, we train a~NER model%
\footnote{Note that this is not a general NER model for addresses, dates, amounts, etc., but rather for more specific data-point types: charity addresses, report dates, incomes, etc.}
and then, at the evaluation stage, we use it for the detection of all occurrences of entities in the text being processed.
    \item \textbf{Normalization}: at this stage objects are normalized to the canonical form which we have defined in the \emph{Kleister} datasets. We use almost the same regular expression as during auto-tagging, e.g.~all detected {\entity{report\_date}} occurrences are normalized from: `November 29, 2019’, `11/29/19’ and \noqa{latex-8}`11-29-2019’ into `2019-11-29’.
    \item \textbf{Aggregation}: we produce a~single output from multiple candidates detected by the NER model. In our case, the technique is simple: we return the object with the maximum summarized scores grouped by the normalized forms of the extracted entities.
\end{enumerate}
Certainly, almost each stage of the above process can be done with a wide range of techniques, from regular expressions to more advanced machine learning models and deep neural networks.

\subsubsection{Pipeline based on Flair}
\label{sec:baseline_flair}
The Flair model (based on stacked char-Bi-LSTM language model and GloVe word embeddings~\cite{Pennington14glove:global}) is used as an encoder and Bi-LSTM with a~CRF layer—as an output decoder.\footnote{We used implementation from the Flair library~\cite{akbik-etal-2018-contextual} in version 0.4.5.} Based on many experiments on the NDA and Charity development datasets, we found out the best setup for parameters, which is: $\mathit{learning\, rate}=0.1$, $\mathit{batch\, size}=32$, $\mathit{hidden\, size}=256$, $\mathit{epoch}=30/15$ (NDA/Charity), $\mathit{patience}=3$, $\mathit{anneal\, factor}=0.5$ and with a~CRF layer on the top. Moreover, each document was split into chunks of 300 words with overlapping parts of 15 words. Results from overlapping parts were normalized into one by using the mean of probabilities for each word from both overlapping parts.%

\subsubsection{Pipeline based on BERT/RoBERTa}
\label{sec:baseline_bert}
The BERT/RoBERTa models are fine-tuning approaches based on the Bidirectional Encoder Representations from Transformers language model.\footnote{We used implementation from \noqa{spell-pytorch}\textit{pytorch-transformers}~\cite{pytorch-transformers} library.} We found out the best experimental setup, which is: $\mathit{learning\, rate}=2e-5$, $\mathit{batch\, size}=8$,  $\mathit{epoch}=20$, $\mathit{patience}=2$ after many experiments on the NDA and Charity datasets for both models. Moreover, each document was split into chunks of 510 tokens (plus two special tokens: \code{[CLS]} and \code{[SEP]}) with overlapping parts of 100 tokens. Results from overlapping parts were normalized into one by using the mean of probabilities for each token from both overlapping parts.%

\subsection{PDF processing tools}
\label{sec:pdf_tools}

PDF documents are input for Kleister challenges, from which we must extract text for further processing. Thus, an important role in whole system and final score of the \textit{Pipeline} approach will be accuracy of the tool that we use. Therefore, we checked performance on three PDF\noqa{grammar-CD_NN} processing tools and one combination of them:

\begin{enumerate}
    \item Tesseract~\cite{tesseract} in version \noqa{spell-rc1}\noqa{spell-gb36c}4.1.1-rc1-7-gb36c\footnote{We ran it with \code{-{}-oem 2 -l eng -{}-dpi 300} flags (meaning both new and old OCR engines were used simultaneously, and language and pixel density were forced for better results)}. This is the most popular free OCR engine currently available.\noqa{latex-8}
    \item Textract with API version from March 1, 2020\footnote{\url{https://aws.amazon.com/textract/}}. One of the most recognizable OCR tools and an open-source competitor of Tesseract.
    \item pdf2djvu/djvu2hocr in version 0.9.8\footnote{\url{http://jwilk.net/software/pdf2djvu}, \url{https://github.com/jwilk/ocrodjvu}} (later we will call that method \textit{pdf2djvu}). Free tool for object extraction (and text extraction) from born-digital PDF files.
    \item pdf2djvu+Tesseract is a combination of pdf2djvu/djvu2hocr and Tesseract. Documents are processed with both tools, by default we take the text from pdf2djvu/djvu2hocr, unless the text returned by Tesseract is 1000 characters longer. This is a simple and efficient way to merge PDF processing  solutions for extracting text from scans and born-digital PDF files.
\end{enumerate}

\subsection{Results}\label{sec:baseline_results}

\begin{table*}[ht!]
\caption{Performance of different PDF processing tools checked on \emph{Kleister} challenges test-sets. All results are f-scores over 3 runs with standard deviation. (*) pdf2djvu does not work on scans, so we have empty 54/24/21 documents in train/dev/test\noqa{spell-dev} sets.\label{tab:ocr_baselines}}
\vspace{1mm}
\centering
\small{
\begin{tabular}{p{3.6cm}p{2.5cm}|p{2.5cm}|p{2.5cm}}\noqa{grammar-UNIT_SPACE}\noqa{latex-44}
\toprule
\multicolumn{4}{c}{\emph{Kleister-NDA} dataset (born-digital PDF files)}\\
\midrule
PDF processing tool & \multicolumn{1}{l}{\textbf{Flair}} & \multicolumn{1}{l}{\textbf{BERT-base}} & \multicolumn{1}{l}{\textbf{RoBERTa-base}}\\
\midrule
pdf2djvu & \textbf{77.70 $\pm$ 0.01} & \textbf{72.17 $\pm$ 1.07} & \textbf{77.07 $\pm$ 1.61} \\
Tesseract & 75.17 $\pm$ 0.12 & 68.30 $\pm$ 0.41 & 74.20 $\pm$ 1.90 \\
Textract & 75.63 $\pm$ 0.17 & 70.43 $\pm$ 1.11 & 76.20 $\pm$ 0.64 \\
\toprule
\multicolumn{4}{c}{\emph{Kleister-Charity} dataset (mixture of born-digital and scans PDF files)} \\
\midrule
pdf2djvu (*) & 71.80 $\pm$ 0.35 & 67.53 $\pm$ 0.71 & 72.50 $\pm$ 0.49 \\
Tesseract & 72.87 $\pm$ 0.81 & 71.37 $\pm$ 1.25 & 75.70 $\pm$ 0.57 \\
Textract & \textbf{80.10 $\pm$ 0.35} & \textbf{73.30 $\pm$ 0.43} & \textbf{79.87 $\pm$ 0.65} \\
pdf2djvu$+$Tesseract & 74.00 $\pm$ 1.28 & 70.47 $\pm$ 0.26 & 75.63 $\pm$ 0.68 \\
\bottomrule
\end{tabular}
}
\end{table*}

\subsubsection{Text extraction methods}
\label{sec:ocr_engine_results}

Results of the performance of different PDF processing tools are presented in Table~\ref{tab:ocr_baselines}. The general conclusion is that by using software with the best text extraction methods we could achieve much better results, but we still cannot resolve all problems connected to information extraction task.

The best tool for born-digital documents in \emph{Kleister-NDA} challenge for all models is a \entity{pdf2djvu} tool. We should expect it, thus that documents are without any text errors normally caused by using OCR engines which are not perfect yet. For \emph{Kleister-Charity} challenge the best tool is Textract with huge advantage on all models, especially flair based. In the next sub-section, we describe in detail baseline results based on the most accurate PDF processing tool for each Kleister task.

\subsubsection{NER models}
\label{sec:ner_results}

The results for the two Kleister datasets obtained with the \textit{Pipeline} method based on Flair, BERT and RoBERTa models are shown in Table~\ref{tab:baselines}. The differences in F-score between Flair and RoBERTa models are not substantial in both challenges. Moreover, the RoBERTa model is much better as far as amounts ({\entity{income\_annually}} and {\entity{spending\_annually}} entities in Charity dataset) and organization names ({\entity{party}} entity in NDA dataset and {\entity{charity\_name}} entity in Charity dataset) are concerned.

The most challenging problems for all models are types related to the business reasoning (e.g.~in NDA {\entity{term}} or {\entity{address\_\_*}}), the visual features (e.g.~in Charity {\entity{income}} and {\entity{spending}}) and finally hard normalization (e.g.~in Charity {\entity{charity\_name}}). We can also observe that the entities appearing in the sequential contexts achieve a~higher F-score. Moreover, after analyzing the model results, we prepared a list of common problems in models, which we grouped into specific problem categories (see Table~\ref{tab:common_problems}).

\begin{table*}[ht!]
\caption{Results of our baselines for \emph{Kleister} challenges test-sets. Results for all models are f-scores over 3 runs with standard deviation. Human baseline is a percentage of annotators agreements for 100 random documents.\label{tab:baselines}}
\vspace{1mm}
\centering
\small{
\begin{tabular}{p{3.9cm}p{2.5cm}|p{2.5cm}|p{2.5cm}|p{2.5cm}}\noqa{grammar-UNIT_SPACE}\noqa{latex-44}
\toprule
\multicolumn{5}{c}{\emph{Kleister-NDA} dataset (pdf2djvu)} \\
\midrule
Entity type & \multicolumn{1}{l}{\textbf{Flair}} & \multicolumn{1}{l}{\textbf{BERT-base}} & \multicolumn{1}{l}{\textbf{RoBERTa-base}} &\multicolumn{1}{l}{\textbf{Human baseline}}\\
\midrule
\entity{effective\_date} & \textbf{82.03 $\pm$ 1.72} & 78.90 $\pm$ 0.86 & 78.53 $\pm$ 2.70 & 100 \% \\
\entity{party} & 70.13 $\pm$ 0.11 & 68.50 $\pm$ 1.92 & \textbf{78.47 $\pm$ 0.58} & 98 \% \\
\entity{jurisdiction} & \textbf{93.80 $\pm$ 0.42} & 92.07 $\pm$ 0.61 & 92.87 $\pm$ 0.90 & 100 \% \\
\entity{term} & \textbf{60.82 $\pm$ 26.7} & 40.23 $\pm$ 1.01 & 42.03 $\pm$ 4.41 & 95 \% \\
\midrule
ALL & \textbf{77.70 $\pm$ 0.01} & 72.17 $\pm$ 1.07 & 77.07 $\pm$ 1.61 & 97.86\% \\
\toprule
\multicolumn{5}{c}{\emph{Kleister-Charity} dataset (Textract)} \\
\midrule
\entity{address\_\_post\_town} & \textbf{83.30 $\pm$ 3.81} & 73.57 $\pm$ 2.49 & 77.70 $\pm$ 1.27 & 98 \% \\
\entity{address\_\_postcode} & \textbf{82.63 $\pm$ 0.54} & 79.00 $\pm$ 0.65 & 82.57 $\pm$ 0.56 & 100 \% \\
\entity{address\_\_street\_line} & \textbf{68.17 $\pm$ 4.36} & 61.33 $\pm$ 2.74 & 63.80 $\pm$ 3.27  & 96 \% \\
\entity{charity\_name}  & 72.40 $\pm$ 0.98 & 73.53 $\pm$ 3.16 & \textbf{76.87 $\pm$ 0.37} & 99 \% \\
\entity{charity\_number} & \textbf{96.73 $\pm$ 0.12} & 96.43 $\pm$ 0.52 & 96.13 $\pm$ 0.45 & 98 \% \\
\entity{income\_annually} & 70.93 $\pm$ 0.43 & 69.97 $\pm$ 1.68 & \textbf{73.20 $\pm$ 0.64} & 97  \% \\
\entity{report\_date} & \textbf{95.67} $\pm$ 0.00 & 94.97 $\pm$ 0.38 & 95.53 $\pm$ 0.12 & 100 \% \\
\entity{spending\_annually} & 68.50 $\pm$ 0.26 & 67.30 $\pm$ 1.36 & \textbf{71.27 $\pm$ 0.84} & 92 \% \\
\midrule
ALL & \textbf{80.10} $\pm$ 0.35 & 77.30 $\pm$ 0.43 & 79.87 $\pm$ 0.65 & 97.45  \% \\
\bottomrule
\end{tabular}
}
\end{table*}

\section{Discussion and Challenges}
\label{sec:discussion}

In Table~\ref{tab:dataset_summary}, we gathered the most important information about open datasets, especially we outlined the difference between our datasets and other sets. Additionally, we prepared descriptions of problems related to Kleister tasks (see Table~\ref{tab:common_problems}). Thus, the \emph{Kleister} datasets appear to be more focused on the real life examples, where layout, document-level context, OCR quality, business logic and normalization problems need to be resolved for obtaining good results.

Summing up, the proposed datasets are useful for testing real life applications to solve the challenge of the robotic process automation tackled by machine learning techniques.

\begin{table}[htbp]
\setlength{\tabcolsep}{3pt}
\caption{Common problems in \emph{Kleister} datasets with examples.\label{tab:common_problems}}
\centering
\begin{tabular}{p{1,2cm}p{13,3cm}}\noqa{grammar-UNIT_SPACE}
\toprule
 \multicolumn{2}{c}{\small{\textbf{Normalization}: Differences in the way entities are given in expected values and documents.}}\\
 \midrule
 \small{\emph{NDA}} & \small{\entity{effective\_date}: October 24, 2012, 10/24/12 or 24th day of October, 2012\noqa{grammar-COMMA_AFTER_A_MONTH}\noqa{proselint-dates_times.dates} \newline \entity{term}: 2 years, 24 months, two (2) years, two years or second anniversary} \\
 \small{\emph{Charity}} & \small{\entity{charity\_name}: 1. Ltd vs Limited: King’s Schools \noqa{spell-Taunton}Taunton LTD [expected] vs King’s Schools Taunton Limited [document]; 2. The vs non-The: The League of Friends of the \noqa{spell-Exmouth}Exmouth Hospital [expected] vs League of Friends of Exmouth Hospital [document]} \\
 \midrule
 \multicolumn{2}{c}{\small{\textbf{Layout}: understand complex layout properly}} \\
 \midrule
  \small{\emph{NDA}} & \small{\entity{all entities}: four types of layout: 1. Simple layout (one column), 2. Simple layout (two columns), 3. E-mail, 4. Plain text. See Fig.~\ref{fig:NDA_layout} in the Appendix.}\\
 \small{\emph{Charity}} & \small{\small{\entity{all entities}: three types of layouts: 1. Simple document, 2. Report with tables, graphic elements and pictures, 3. Form. See Fig.~\ref{fig:charity_layout}.}}\\
\midrule
 \multicolumn{2}{c}{\small{\textbf{Document-level context}: understand document as a whole}} \\
\midrule
 \small{\emph{NDA}} & \small{\entity{term}: The term informs about the duration of the contract. Information on this is generally found in the “Term” chapter. However, this section may also include other periods of validity of certain provisions of the contract. \newline
 Example: “Term. This Agreement will be effective for a period of one (1) year after the Effective Date. The restrictions on use and disclosure of the Discloser’s Confidential Information by the Recipient shall survive any expiration or termination of this Agreement and shall continue in full force and effect for a period of five (5) years thereafter.”} \\
 \small{\emph{Charity}} & \small{\entity{income\_annually}, \entity{spending\_annually}: Co-occurrence of exact and rounded values in one document. See Fig.~\ref{fig:charity_diff_values} in the Appendix}\noqa{grammar-PUNCTUATION_PARAGRAPH_END} \\
\midrule
\multicolumn{2}{c}{\small{\textbf{Business logic}: apply some rules in a case of ambiguity}} \\
\midrule
 \small{\emph{NDA}} & \small{\entity{term}: Co-occurrence of two terms in one document. In such a case, the one constituting the duration of the renewed contract was considered inappropriate. \newline
 Example: ‘Term; Termination. The term of the employment agreement set forth in this shall be for a period commencing at the Effective Date and continuing for three (3) years thereafter (the “Scheduled Term”). Following the Scheduled Term, the Agreement shall automatically renew for successive one-year terms (each a “Renewal Term”).’}\\
 \small{\emph{Charity}} & \small{\entity{address\_\_*}: Co-occurrence of different addresses (e.g. Principal address, Registered office, Administrative address, etc.) next to each other in one document, or the lack of a clear identification of the charity's address. In such a case the Registered address was considered to be the main one. See Fig.~\ref{fig:datasets_problems}.}\\
 \midrule
 \multicolumn{2}{c}{\small{\textbf{OCR quality}: process scan documents}} \\
 \midrule
 \small{\emph{NDA}} & \small{N/A — born-digital documents.}\\
 \small{\emph{Charity}} & \small{\entity{all entities}: Handwriting in the document, pages upside down or poor scan quality.}\\
 \bottomrule
\end{tabular}
\end{table}

As described above, working with the proposed datasets can be compared to challenges dealing with Information Retrieval and Natural Language Understanding, including challenges related to page layout understanding (i.e.~tables, rich graphics, etc.). To solve these challenges, we presented the \textit{Pipeline} approach that will help to deal with specific problems.

Most of these stages are described in the process of building baselines and are shown in Fig.~\ref{fig:data-baselines}.

Using the presented challenges we are also able to study the impact of each stage of the full process on the final results. It is useful in the production environment where we can have a~baseline, and then we can assess what should be done with the highest priority to improve final results.

\section{Conclusions}
\label{sec:conclusions}

\emph{Kleister} datasets have been prepared to challenge the business usability of Information Extraction models and processes. In this article, we described in detail how they were prepared (i.e.~manually or automatically --- for \emph{Kleister-NDA} and \emph{Kleister-Charity} respectively). Due to their multi-modal nature, we had to face various problems and needed to develop methods to improve the quality of data sets.

We consider our datasets and tasks will help the community to extend the understanding of documents with substantial length, various reasoning problems, complex layouts and OCR quality problems. Moreover, the community can use our methodology to extend the datasets or prepare similar sets.

In addition, we prepared baseline solutions on the basis of text data generated by different PDF processing tools from the datasets (see Table~\ref{tab:ocr_baselines}). This benchmark shows weakness of the models working on a~pure text (i.e. input is a~sequence of words) without using any visual information.

\bibliography{ms}
\bibliographystyle{acl}

\clearpage
\appendix

\section*{Appendix}

In this supplement we describe more precisely our datasets  and the
annotation processes in Section~\ref{sup:sec:NDA_dataset}
and~Section~\ref{sup:sec:charity_dataset}, respectively.%

\section{NDA Dataset}\label{sup:sec:NDA_dataset}

\subsection{Data Detailed Description}\label{sec:data_description_nda}

The NDA agreements prevent the disclosure of confidential information by one of the parties to a~third party. Such agreements, even in oral form, are often found in everyday life (e.g. in the patient-doctor relationship). In business, they usually have a~written form, signed by a~representative of the legal profession and another person (legal or natural). In our database, we have collected business contracts, but without differentiating them, either by their form (these are both independent contracts and contracts annexed to other contracts), or by the way they were concluded (all contracts were concluded in writing, some of them by e-mail) or because of the number of parties (the dataset contains unilateral, bilateral and multilateral agreements).

The NDAs can take various forms (contract attachments, emails, etc.), but they all generally have a~similar structure. First, the circumstances of the contract are determined, i.e. the parties to the contract are presented and the date from which the contract becomes effective is provided. Then they usually contain the following elements:

\begin{itemize}
\item a~definition of confidential information, including exceptions to this definition;
\item description of the disclosure procedure (also during court and administrative proceedings);
\item procedures related to non-compliance with confidentiality obligations;
\item term of the contract (termination date);
\item the period during which the information remains confidential (confidential period);
\item information about the jurisdiction to which the contract is subject;
\item information about the possibility of making legally binding copies of the contract;
\item due to the fact that confidential information can be used to recruit new employees or contractors of one party by another, the NDA often also includes non-compete clauses in force for a~certain period of time.
\end{itemize}

\begin{figure*}[ht!] %
\begin{center}
\includegraphics[width=15cm]{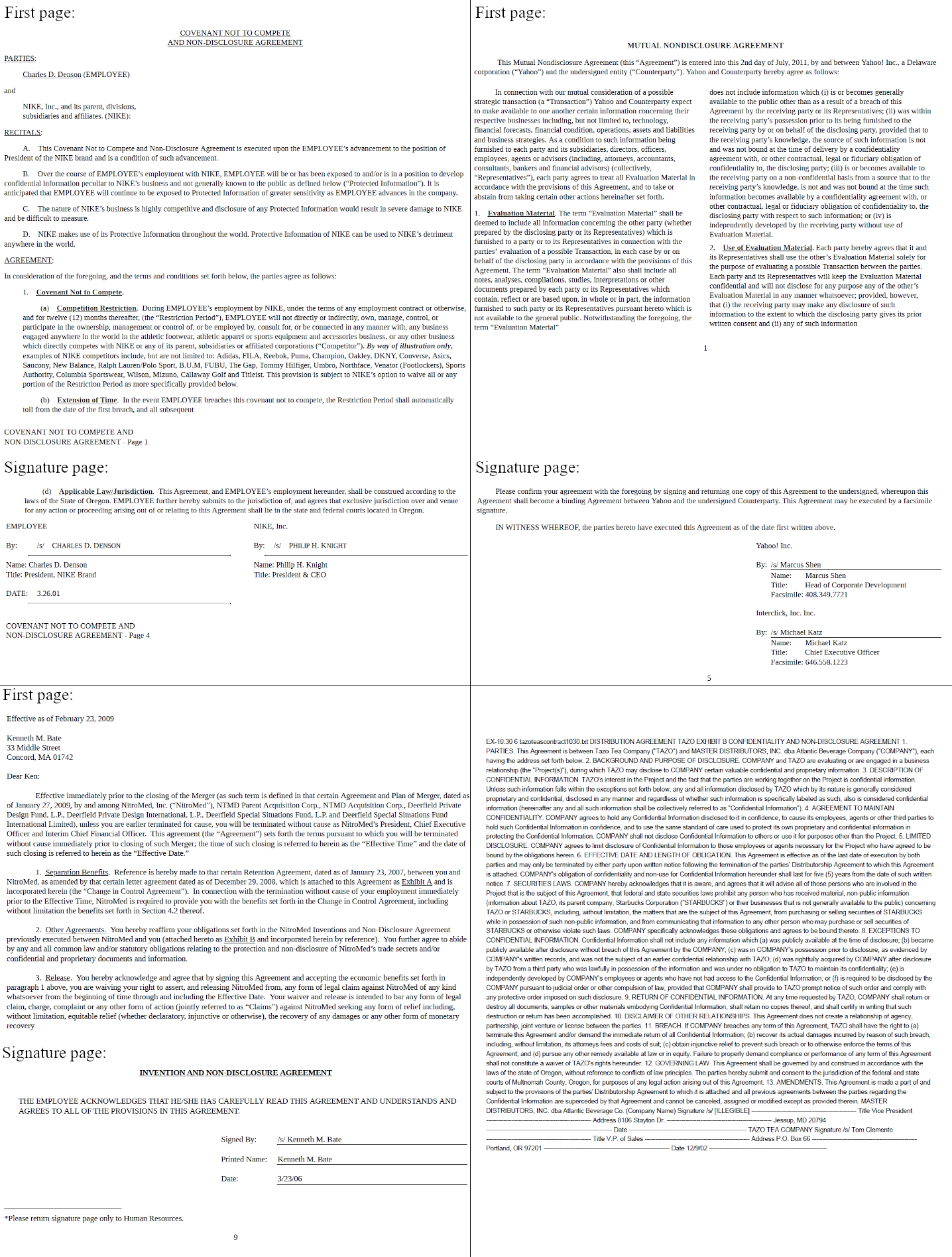}
\caption{Four different examples of layouts from the \emph{NDA-charity} dataset.\label{fig:NDA_layout}}
\end{center}
\end{figure*}

\subsection{Data Collection Method}\label{sec:data_collection_nda}

During the collection of the NDAs, we focused on contracts concluded by public companies in the United States. All public companies (i.e. those with shareholders) in the US are supervised by the United States Securities and Exchange Commission (SEC). Companies are required to submit a~number of reports and forms, the attachments of which are often contracts concluded by these companies, including NDAs. This is done through the Electronic Data Gathering, Analysis and Retrieval system (EDGAR), which is also a~public database of these documents (these documents must be made public)\footnote{\url{https://www.sec.gov/edgar.shtml}}. As a~result, EDGAR is a~huge NDA base. Unfortunately, NDAs are usually attachments to other contracts or forms submitted to EDGAR, as a~result of which it is not possible to simply aggregate them from this database. Thus, the process of gathering the dataset had to be manual, with a~weak model supervision.

The NDAs were collected with the help of the Google search engine. Two collections were created—the first contained 170 contracts and the second 330 contracts, except that 117 duplicates were found, so that ultimately the dataset counted a~total of 383 documents. After the first tests on the already annotated dataset, it turned out that machine learning models achieve quite poor results for information on jurisdiction. Analysis of the dataset showed that this was due to the under-representation of documents that were prepared in accordance with non-US law (e.g. China, India or Israel). Since no more such documents were obtained, the 68 previously obtained ones were removed from the dataset, which reduced it to 315 documents. In the next step, the collection was supplemented with an additional 127 documents consistent with the others in terms of applicable law (i.e. US law).

The original files were HTML documents, but they were transformed into PDF files to keep processing simple and similar to how other datasets were created. Transformation was made using the \texttt{puppeteer} library, which in turn used the “Print to PDF” functionality present in the \texttt{Chrome} web browser. Subsequently, the transformed PDFs were processed with the Tesseract OCR engine.

\subsection{Annotation Procedure}\label{sec:data_annotation_nda}

The whole dataset was annotated in two ways. Its first part, i.e. 315~documents, was annotated by linguists, except that only selected contexts, preselected by an in-house system based on semantic similarity, were taken into account (to make the annotation easier and faster). The second, i.e. 127 documents, was entirely annotated by hand. When preparing the dataset, we wanted to find out if the semantic similarity methods could be used to limit the time it would take to perform annotation procedures (this solution saved about 50\% of the time compared to fully manual annotation).

The\noqa{grammar-ENGLISH_WORD_REPEAT_BEGINNING_RULE} annotation of the dataset consisted of listing the extracted entities. The entities themselves may appear repeatedly in the document, but this did not matter for the annotation procedure (contrary to NER, we are not interested in the exact location(s) of an entity). The following entities have been normalized according to standards adopted by us: (a) {\entity{parties}} --- commas have been removed before acronyms referring to organization types, and the format has been unified, e.g.~\textit{LHA\noqa{spell-LHA} LONDON LTD}; (b) {\entity{effective date}} --- the format has been standardized according to ISO~8601, \noqa{spell-YYYY}i.e.~YYYY-MM-DD; \noqa{proselint-typography.symbols.copyright}(c) {\entity{terms}} --- standardized to the following format: number of units followed by a~unit, e.g.~\textit{2 years}; (d) {\entity{jurisdiction}} and {\entity{counterparts}} did not require standardization. Then the annotations were checked by the super-annotator on 45 random documents (10\% of the whole dataset). All the super-annotated entities were correct and did not need to be changed.

\section{Charity Dataset}
\label{sup:sec:charity_dataset}

\subsection{Data Detailed Description}
\label{sup:sec:data_description}

There is no rule about how such a~charity report should look. Therefore, some take the form of reports richly illustrated with photos and charts, where financial information constitutes a~small part of the entire report, while others have only a~few pages, where only basic data on revenues and expenses in a~given calendar year are given (see Figure~\ref{fig:charity-web-page}). However, each of these reports should contain at least the following information (although there may be exceptions to this rule):
\begin{itemize}
\item organization's address, name and number;
\item the date of submission of the report;
\item total income in the reporting year;
\item total expenditure in the reporting year.
\end{itemize}

\subsection{Data Collection Method}
\label{sec:data_collection_charity}

\begin{figure*}[h!t]
\begin{tabular}{cc}
 \includegraphics[width=7.5cm]{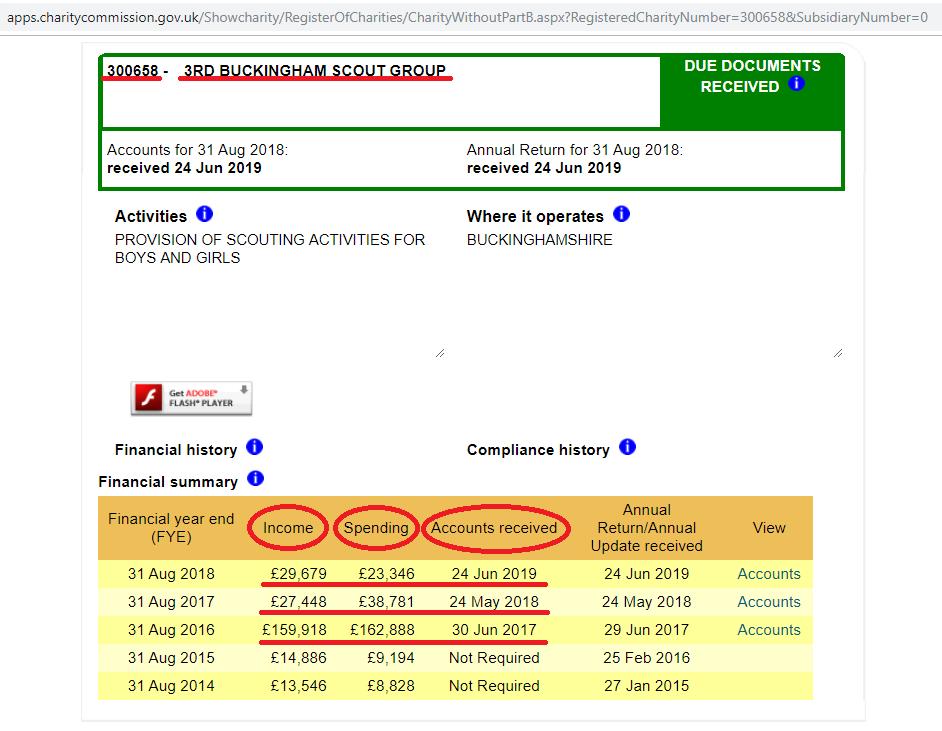}    &
 \includegraphics[width=7.5cm]{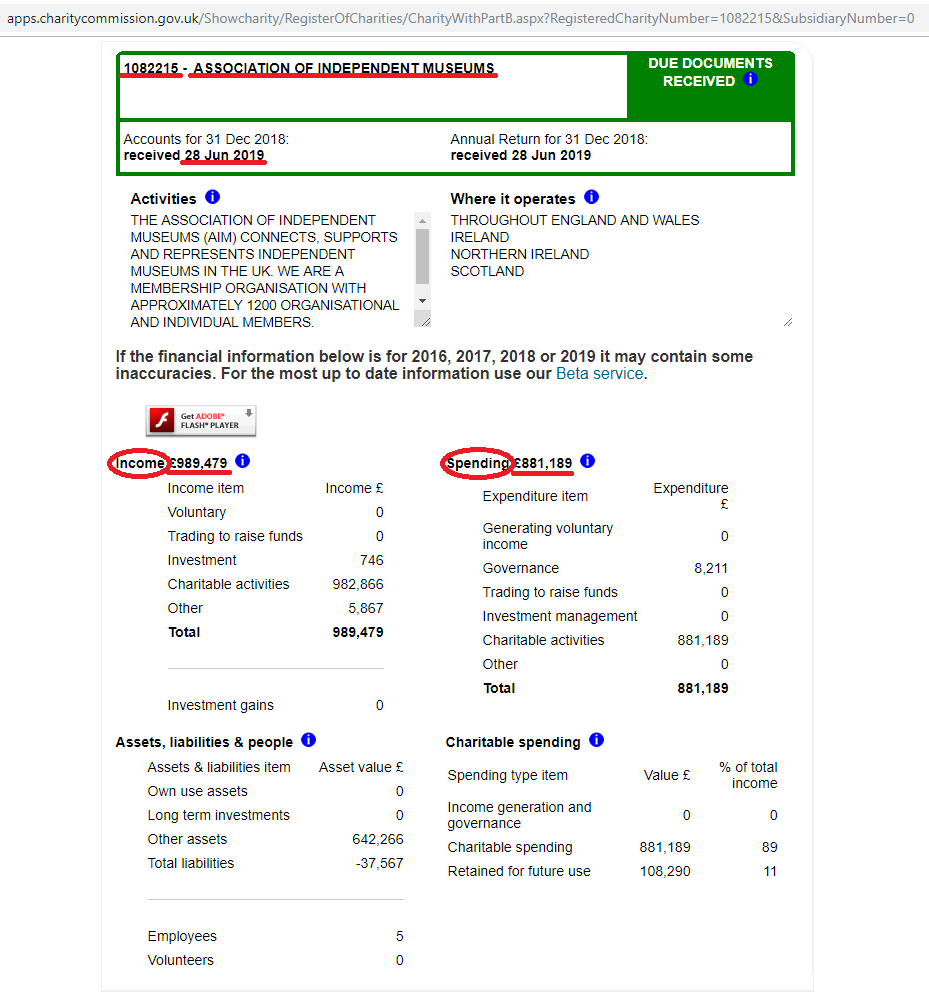}
\end{tabular}
\caption{Organization's page on the Charity Commission's website (left: organization whose annual income is between 25k\noqa{spell-25k} and 500k\noqa{spell-500k} GBP, right: over 500k). Information on the website has a~different layout, and within documents there is also the case. Entities are underlined in red and names of entities are circled.\label{fig:charity-web-page}}
\end{figure*}

The decision to create a~dataset from the financial reports of British charities was driven by the following goal: to find a~publicly available collection of English-language and multi-page documents on the Internet, which would be accompanied by easy-to-extract information about data contained in these documents (e.g. as a~separate XML file or a~table on a~website). We decided that the database of financial reports of British charity organizations would be the best of all the options considered. It is not just that the Charity Commission website actually has a~database of all the charity organizations registered in England and Wales, but also that each of these organizations has a~separate subpage on the Commission's website and it is easy to find the most important information about them (see Fig.~\ref{fig:charity-web-page}):
\begin{itemize}
\item Charity's name and number;
\item main activities;
\item current address parts (post town, postcode and street line);\noqa{grammar-EN_COMPOUNDS}
\item a~list of the current trustees of the organization;
\item basic financial data for the past year, i.e.~income and expenditure (these data are more detailed in the case of organizations with revenues of over 500,000 GBP a~year);
\item the date of submission of the report.
\end{itemize}
This information partly overlaps with what the reports actually contain (although it might happen that some entities are not to be found in the reports, e.g. a~list of trustees is given on the website, but it does not have to be included in the report). For this reason, we decided to extract only those entities which also appear in the form of a~brief description on the website.

The reports can be found on the website as PDF files (but this does not apply to organizations with income below 25,000 GBP a~year, as they are required to submit a~condoned financial report). Therefore, the information available on the website and the documents attached to it made the database of these documents perfectly fit the objectives outlined above. In this way, 3414 documents were obtained.

During the analysis of the documents, it turned out that several reports are in Welsh. As we are interested in the English language only, all documents in other languages were found and removed from the collection. In addition, documents, that contained reports for more than one organization, were handwritten, or the quality of their OCR was low, were deleted. As a~result, the collection has 2778 documents.

\subsection{Annotation Procedure}
\label{sec:data_annotation_charity}

There was no need to manually annotate documents, because basic information about the reporting organizations could be obtained directly from the website where these documents were located.

Only a~random sample of 100 documents was manually checked (see Table~\ref{tab:charity_data_comparison}). The permissible error limit for a~given entity was set at 15\%. These results were exceeded for \entity{charity name} (18\% of errors and minor differences) and for \entity{charity address} (76\% of errors and minor differences). However, as a~result of detailed analysis, it turned out that there are few erroneous entities (respectively 5\% and 9\%), while the rest is rather due to differences in the way the data is presented on the page and in the document. These minor differences have been corrected manually and automatically, as described below.

\begin{table}[htbp]
\caption{Comparison of data on the Charity Commission's website and in charity reports.\label{tab:charity_data_comparison}}
\centering
\begin{tabular}{p{3.5cm}p{1,2cm}p{1,5cm}p{0,8cm}}\noqa{grammar-UNIT_SPACE}
\toprule
 Entities & Correct [\%] & Minor differences [\%] & Error [\%]\\
 \midrule
 \entity{charity\_name} & 82 & 13 & 5 \\
 \entity{charity\_address} & 24 & 67 & 9 \\
 \entity{charity\_number} & 98 & 0 & 2 \\
 \entity{report\_date} & 99 & 0 & 1 \\
 \entity{income\_annually} & 86 & 3\footnote{Including two cases of non-rounding of the amount and one filling in the amount in USD instead of GBP.} & 11 \\
 \entity{spending\_annually} & 86 & 3\footnote{As above.} & 11 \\
 \bottomrule
\end{tabular}
\end{table}

Hence, the charity's name on the website and in the documents could be noted once with the term \textit{Limited} (shortened to \textit{LTD}), and once not. This problem was eliminated by the manual annotation of all documents in which the name of the charity organization co-occurred with the word \textit{Limited} or \textit{LTD}. As a~result, 366~documents were analyzed manually in this way.

In the case of the charity's address the most problematic were the names of counties, districts as well as the names of towns and cities, which were once specified on the website, but not in the documents, other times—the other way round. This problem was solved by splitting address data into the three separate entities that we considered the most important—postcode, postal town name and street or road name. The postal code was used as the key element of the address, on the basis of which the city name and street name could be determined\footnote{Postal codes in the UK were aggregated from a~website: \url{streetlist.co.uk}}.

\begin{figure*}[ht!]
\begin{center}
\includegraphics[width=16cm]{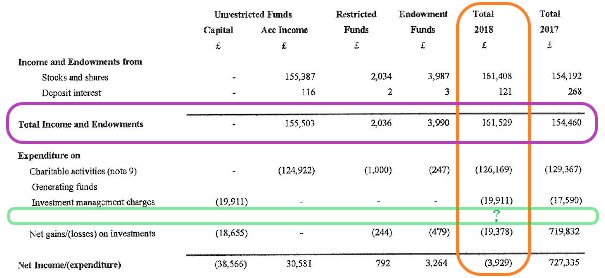}
\caption{No value for \entity{spending\_annually}.\label{fig:charity_no_spend}}
\end{center}
\end{figure*}

\begin{figure*}[ht!]
\begin{center}
\includegraphics[width=16cm]{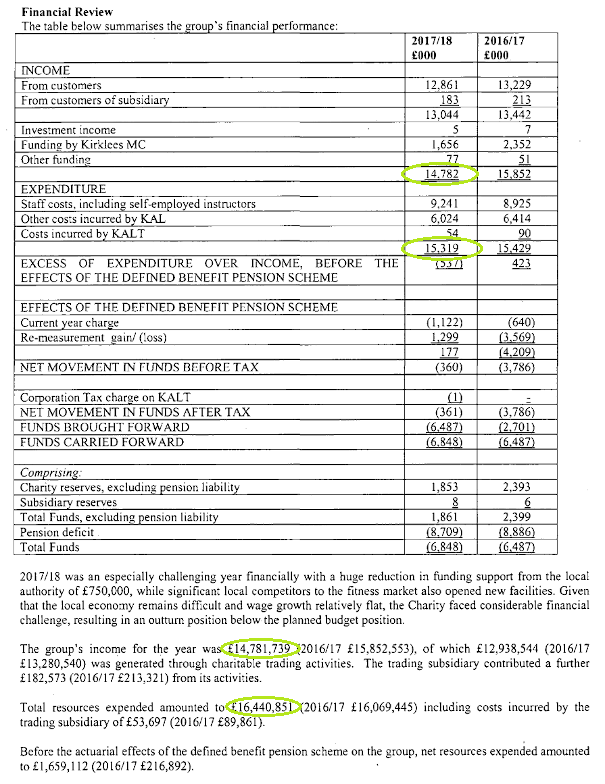}
\caption{Different values for \entity{income\_annually} and \entity{spending\_annually}.\label{fig:charity_diff_values}}
\end{center}
\end{figure*}

Other problems show Fig.~\ref{fig:charity_no_spend} and~\ref{fig:charity_diff_values}. On the first of them we have two different values for \entity{income\_annually} and \entity{spending\_annually}, because the values in the table are rounded and in the text are accurate. In the second picture there is no total for all expenses, so we can not extract the value for \entity{spending\_annually}.

\end{document}